\documentclass[12pt,twocolumn]{article}
\pdfoutput = 1

\usepackage[utf8]{inputenc}
\usepackage[english]{babel}

\usepackage[letterpaper, hmargin={2.5cm,2cm},vmargin={2cm,2.5cm}]{geometry}
\usepackage{times}

\usepackage{standalone, tikz, pgf}
\usetikzlibrary{plotmarks}
\usepackage[caption=false,font=footnotesize]{subfig}
\usepackage{url}
\usepackage[hidelinks]{hyperref}

\usepackage{amsmath}
\usepackage{amssymb}
\usepackage{amsbsy}
\usepackage{amsthm}
\usepackage{bm}

\usepackage{appendix}
\newcommand{\mi}{\mathrm{i}}
\newcommand{\cis}{\mathrm{cis}}
\newcommand{\lex}{>_{\mathrm{lex}}}

\newtheorem{proposition}{Proposition}
\newtheorem{theorem}{Theorem}
\newtheorem{corollary}{Corollary}

\title{Solving the Forward Position Problem of an In-Parallel Planar Manipulator in the Gauss Plane}
\author{Sureyya Sahin\\
  \texttt{sahin508@gmail.com}}
\date{}

\begin{document}
\maketitle
\begin{abstract}
  \textit{We study determining the posture of an in-parallel planar manipulator, which has three connectors composed of revolute, prismatic and revolute joints, from specified active joint variables. We construct an ideal in the field of complex numbers, and we introduce self inversive polynomials. We provide results for an in-parallel planar manipulator, which has a base and moving platform in right triangular shape. Using Sage computer algebra system,  we compute its Groebner bases. We illustrate that the single variable polynomials obtained from the Groebner bases are self reciprocal.}
\end{abstract}
\section{Introduction}
In-parallel planar manipulators bring advantages to their application fields since they provide good load carrying capacity and accuracy; furthermore, these manipulators exhibit a simple kinematic structure. An in-parallel manipulator is composed of connectors as defined in Duffy \cite{duffy_statics_1996}, which are serial identical kinematic chains, a fixed base, and a moving platform. Each of the connectors, acting between the base and the moving platform, has at most one actuator. 

While the kinematic structure of an in-parallel planar manipulator is relatively simple, a forward position analysis of such manipulators, which involves determining the posture of the moving platform for given active joint displacements, can be difficult. There are studies on solving this problem by using parametrization in the field of real numbers ($\mathbb{R}$). Nielsen and Roth \cite{nielsen_roth_1999} reported developments related to computational and numerical solution techniques of this problem including polynomial continuation and Groebner bases theory; furthermore, they presented application of the elimination methods to serial and parallel manipulators. As opposed to commonly used methods in solving the forward position analysis problem of planar manipulators as reported in \cite{nielsen_roth_1999}, which are parametrized in the field of real numbers ($\mathbb{R}$), we obtain the solution of the equations in the algebraically closed field of complex numbers ($\mathbb{C}$). A few studies on using complex number methods are available in the literature. For example, Shigley et al. \cite{shigley_theory_1995} used complex number methods to study the position analysis of closed-loop mechanisms. However, they treated complex numbers as an algebraic manipulation tool; therefore, they obtained solutions by utilizing the real and imaginary components of the resulting equations in $\mathbb{R}$. Another study available is by Wampler\cite{wampler_solving_2001}, who utilized numerical and matrix methods to solve forward position analysis problem of a closed-loop mechanism. The numerical method in \cite{wampler_solving_2001}, which was formulated in the Gauss plane led to solution in the field of complex numbers $\mathbb{C}$. Nevertheless, we attempt to solve the forward position analysis problem of an in-parallel planar manipulator in $\mathbb{C}$ by using Groebner bases theory, which is one of the fundamental computational tools in algebraic geometry. The method in this study leads us to self reciprocal polynomials, which typically appear in problems involving unit circle. These polynomials do not seem to be reported in kinematics literature before, yet there are studies in mathematics. For example, Diaz-Barrero and Egozcue \cite{diaz2004characterization} studied a methodology for constructing self inversive polynomials. We do not construct such polynomials in this study, but we obtain them from the algorithm for computing the Groebner basis.

The in-parallel planar parallel manipulator, which is used in this study has three connectors composed of revolute, prismatic, and revolute (3RPR) joints. In Section \ref{sec:inversion_unit_circle}, we introduce inversion in the unit circle and derive self inversive polynomials, which possess a point and its inverse as roots. In Section \ref{sec:solution}, we discuss variety of an ideal, then solve the forward position problem of the manipulator by using the elimination ideals of its Groebner basis. We utilize the position equations of the 3RPR manipulator in the Gauss plane, which was derived in Sahin \cite{2015arXiv150801733S}, in Section \ref{sec:solution}. Computer algebra systems are widely used in computing Groebner bases; thus, we use Sage mathematical software \cite{sage} for computations in the field of Gaussian Rational numbers ($\mathbb{Q}(\mi)$). We perform simulations which give two and four solutions--the 3RPR manipulator may have up to six solutions \cite{duffy_statics_1996}--for the forward position analysis problem; additionally, we illustrate the inversive nature of the solutions in the unit circle. We draw the assembly configurations to illustrate the possible postures of the manipulator for each simulations.


\section{Inversion of Points in the Unit Circle}\label{sec:inversion_unit_circle}
This section includes background in inversive points and inversion in the unit circle as well as self reciprocal polynomials and their relation to inversive points.

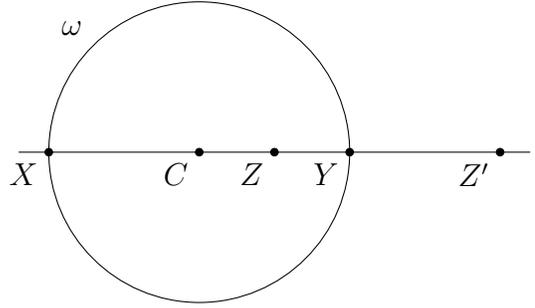
\begin{figure}[!t]
	\begin{center}
			\begin{tikzpicture}[scale=2]
	\draw (-1.2,0) -- (2.2,0);
	\draw (0,0) circle (1) node [above left=40pt] {$\omega$};
	\draw [fill=black] (0,0) circle (0.025) node [below left] {$C$};
	\draw [fill=black] (0.5,0) circle (0.025) node [below left] {$Z$};
	\draw [fill=black] (2,0) circle (0.025) node [below left] {$Z'$};
	\draw [fill=black] (-1,0) circle (0.025) node [below left] {$X$};
	\draw [fill=black] (1,0) circle (0.025) node [below left] {$Y$};
	\end{tikzpicture}
	\end{center}
	\caption{Power of a point $Z$ with respect to the unit circle.}
	\label{fig:power_circle}
\end{figure}

To begin with, we refer to Fig. \ref{fig:power_circle}, where a circle $\omega$ centered at $C$ with radius $r$ and a point $Z$ different from $C$ are given. We find a point $Z'$ on the ray through $CZ$ such that $\boldsymbol(CZ)\cdot\boldsymbol(CZ')=r^2$; thus, the point $Z'$ is the inverse of $Z$ in the circle $\omega$ as defined in Dodge \cite{dodge_euclidean_2004}. If we are given a point $Z$ whose inverse in $\omega$ is $Z'$, then the inverse of $Z'$ is $Z$. Therefore, inversion is cyclic and has period two.

We represent the inversion in the Gauss plane by using the equation $|z-c||z'-c|=r^2$, $z'$ being the affix of the point $Z'$ in the Gauss plane. Consequently, we provide the following theorem the proof of which is available in \cite{dodge_euclidean_2004}.
\begin{theorem}\label{thm:inv_circle}
	Inversion in a circle with center $C$ and radius $r$ has the complex representation
	\begin{equation*}
	(z'-c)(\overline{z}-\overline{c})=r^2
	\end{equation*}
\end{theorem}

If we replace the center $C$ of the circle $\omega$ in Fig.3 to coincide with the origin $O$, then we can use Theorem \ref{thm:inv_circle} to obtain inversion in the unit circle. Consequently, substituting $c=0$, and the radius $r=1$ into the equation given in Theorem \ref{thm:inv_circle}, we obtain the Corollary \ref{cor:inv}.
\begin{corollary}\label{cor:inv}
	Inversion in the unit circle $\mathbb{S}$ has the complex representation
	\begin{equation}\label{eq:inv-S}
	z'=\frac{1}{\overline{z}}=\frac{z}{|z|^2}
	\end{equation}
\end{corollary}
Hence, we can calculate the inverse of a point $z$ in $\mathbb{S}$ by using Eq. \ref{eq:inv-S}. For example, if we pick a point $Z$ on $\mathbb{S}$, then its inverse is $z'=z/|z|^2=z$. Thus, points on $\mathbb{S}$ are invariant. Nevertheless, if we pick a point outside (inside) $\mathbb{S}$, then we will obtain the image inside (outside) the circle. Therefore, inversion turns $\mathbb{S}$ inside-out, as referred to by Coxeter and Greitzer \cite{coxeter_greitzer_1967}. Thus, we can think of the Gauss plane as composed of three partitions, the unit circle, inside and outside by using inversion in the unit circle.

We can use inversion to investigate the roots of a polynomial in a single variable, whose coefficients are defined in $\mathbb{C}$. Thus, we pick $z$ as the root of such a polynomial, which is given by Eq. \ref{eq:gen_pol}.
\begin{equation}\label{eq:gen_pol}
f(z) = c_nz^n+c_{n-1}z^{n-1}+\cdots+c_1z+c_0
\end{equation}
Let us suppose that the inverse $1/\overline{z}$ is also root of the polynomial in Eq. \ref{eq:gen_pol}. Then, we must have
\begin{equation}\label{eq:pol-si-int}
c_n(\frac{1}{\overline{z}})^n+c_{n-1}(\frac{1}{\overline{z}})^{n-1}+\cdots+c_1\frac{1}{\overline{z}}+c_0=0
\end{equation}
If we multiply Eq. \ref{eq:pol-si-int} by $\overline{z}^n$ and take the conjugate of the resulting equation, we obtain
\begin{equation}\label{eq:conj_recip_poly}
f^*(z)=z^n\overline{f(\frac{1}{\overline{z}})}=\sum_{i=0}^n\overline{c}_{n-i}z^i=0
\end{equation}
Thus, we obtain the polynomial $f^*$, which is known as the (conjugate) reciprocal of the polynomial $f$. Consequently, we call $f$ as a self inversive polynomial since $f(z)=f^*(z)$. Furthermore, we must have for each constant of $f$ the formula $c_i=\overline{c}_{n-i}$ with $i=0,\hdots,n$. In summary, if $z$ and its inverse in $\mathbb{S}$ are roots of a polynomial, then we have a self inversive polynomial. The converse of this statement is also true. To illustrate this, we let $z$ be the root of a self inversive polynomial $f$. Hence we write $f^*(z)=0$. So, $z^n\overline{f(1/\overline{z})}=0$. Since $z\neq 0$, we have $\overline{f(1/\overline{z})}=0$; thus, $f(1/\overline{z})=0$. Therefore, we conclude that $1/\overline{z}$ is also a root of the self inversive polynomial.

Yet, we consider a special case of self inversive polynomials. We let the constants of the polynomial $f$ in Eq. \ref{eq:gen_pol} be real numbers and a solution $z$ would lie in the real axis. Then, the Eq. \ref{eq:conj_recip_poly} reduces to
\begin{equation}
\label{eq:reciprocal_poly}
f^+(z)=z^nf(\frac{1}{z})=\sum_{i=0}^nc_{n-i}z^i
\end{equation} 
In this case, we call the polynomial in Eq. \ref{eq:reciprocal_poly}, as defined in Barbeau \cite{barbeau1989polynomials}, the reciprocal of $f$. Analogous to self inversive polynomials, we call the polynomial satisfying $f(z)=z^nf^+(1/z)$ self reciprocal. In this case, the reciprocal $1/z$ of the root $z$ is also a root of $f$.

We refer to Fig. \ref{fig:power_circle} again with the circle being $\mathbb{S}$. Furthermore, we let the line on which $Z$, $Z'$ lie be collinear with the real axis. Denoting the intersection points of $\mathbb{S}$ with the real axis as $X=(-1,0)$ and $Y=(1,0)$, we prove the proposition \ref{prop:harmonic_conj} by using the notion of a harmonic conjugate. As given in Pedoe \cite{pedoe_geometry_1988}, the points $Z$ and $Z'$ are harmonic conjugates with respect to the segment $|\mathbf{XY}|$ if the Eq. \ref{eq:harm_conj} is satisfied.
\begin{equation}\label{eq:harm_conj}
\frac{\mathbf{Z'X}}{\mathbf{Z'Y}}=-\frac{\mathbf{ZX}}{\mathbf{ZY}}
\end{equation}
where the bold variables in the fractions denote directed line segments. Thus, we are ready to prove Proposition \ref{prop:harmonic_conj}.
\begin{proposition}\label{prop:harmonic_conj}
Let $z'=1/z$ be the reciprocal of a real number $z$ obtained from inversion in the unit circle. Then, $Z$ and $Z'$ are harmonic conjugates with respect to the intersection points $X$ and $Y$ of the unit circle with the real axis.
\end{proposition}
\begin{proof}
Referring to Fig. \ref{fig:power_circle}, and using the Eq. \ref{eq:harm_conj}, we find
\begin{equation*}
\frac{\mathbf{Z'X}}{\mathbf{Z'Y}}=\frac{-1-\frac{1}{z}}{1-\frac{1}{z}}=\frac{1+z}{1-z}=-\frac{\mathbf{ZX}}{\mathbf{ZY}}
\end{equation*}
Therefore we conclude that $Z'$ and $Z$ are harmonic conjugates with respect to $|\mathbf{XY}|$.
\end{proof}

Thus, if we have inversion of a point in the unit circle along the real axis, then the pair of points are harmonic conjugates.


\section{Solution of the Position Equations}\label{sec:solution}
We will solve the position equations of the 3RPR manipulator, which are derived in \cite{2015arXiv150801733S}, in this section. These equations of the 3RPR manipulator belong to the set of polynomials with coefficients in $\mathbb{C}$, which we denote as
\[\mathbb{C}[\cis_a,\cis_b,\cis_c,\cis_\alpha,\overline{\cis}_a,\overline{\cis}_b,\overline{\cis}_c,\overline{\cis}_\alpha]\] 
Consequently, we would like to show that the position equations generate an ideal. We recall that for a set of polynomials $g_1,\dots,g_n\in k[x_1,x_2,\dots,x_n]$ with $k$ being the field to which the coefficients of the polynomials belong, and $x_i$ being variables of the polynomial, $\left<g_1,\dots,g_n\right>$ is the ideal of $k[x_1,\dots,x_n]$ generated by $g_i$ is given by Cox et al. \cite{cox_ideals_2007} as:
\begin{equation} \label{eq:idealgen}
\left<g_1,g_2,\dots,g_n\right>=\left\{\sum_{i=1}^nh_ig_i:g_i\in k[x_1,\dots,x_n]\right\}
\end{equation}
We use this definition in Proposition \ref{prop:ideal} to prove that the loop-closure equations and the circular constraints derived in \cite{2015arXiv150801733S} form an ideal.
\begin{proposition}\label{prop:ideal}
The position equations of the 3RPR in-parallel planar manipulator generates an ideal given by $I=\left<f_1,f_2,\dots,f_8\right>$ such that
\begin{eqnarray*}
f_1&=&s_a \cis_a+l_{ab} \cis_\alpha-s_b \cis_b-d_{ab}\\
f_2&=&s_a\cis_a+l_{ac}\cis_\beta\cis_\alpha-s_c\cis_c-d_{ac}\\
f_3&=&s_a\overline{\cis}_a+l_{ab}\overline{\cis}_\alpha-s_b\overline{\cis}_b-\overline{d}_{ab}\\
f_4&=&s_a\overline{\cis}_a+l_{ac}\overline{\cis}_\beta\overline{\cis}_\alpha-s_c\overline{\cis}_c-\overline{d}_{ac}\\
f_5&=&\cis_a\overline{\cis}_a-1\\
f_6&=&\cis_b\overline{\cis}_b-1\\
f_7&=&\cis_c\overline{\cis}_c-1\\
f_8&=&\cis_\alpha\overline{\cis}_\alpha-1
\end{eqnarray*}
in  $\mathbb{C}[\cis_a,\cis_b,\cis_c,\cis_\alpha,\overline{\cis}_a,\overline{\cis}_b,\overline{\cis}_c,\overline{\cis}_\alpha]$.
\end{proposition}
\begin{proof}
We use the definition of the ideal in the proof:
\begin{enumerate}
\item $\left<f_1,f_2,\dots,f_8\right>$ has zero element: We write $\sum_{i=1}^80f_i=0$ for $0\in\mathbb{C}$.
\item $\left<f_1,f_2,\dots,f_8\right>$ is closed under addition: We suppose that $f=\sum_{i=1}^8 p_if_i$, and $g=\sum_{i=1}^8q_if_i$ where
$
  p_i,q_i \in\mathbb{C}[\cis_a,\cis_b,\cis_c,\cis_\alpha,\overline{\cis}_a,\overline{\cis}_a,\overline{\cis}_b,\overline{\cis}_c,\overline{\cis}_\alpha]$
Then, the equation
\begin{equation*}
f+g=\sum_{i=1}^8(p_i+q_i)f_i\in \left<f_1,f_2,\dots,f_8\right>
\end{equation*}
holds.
\item Pick \[h\in\mathbb{C}[\cis_a,\cis_b,\cis_c,\cis_\alpha,\overline{\cis}_a,\overline{\cis}_b,\overline{\cis}_c,\overline{\cis}_\alpha]\]
Then, we observe $hf=\sum_{i=1}^8(hp_i)f_i\in I$.
\end{enumerate}
Since $I$ satisfies the conditions of an ideal, $I$ is an ideal.
\end{proof}

Thus, we construct an ideal from a finite number of polynomial equations--consequently, every ideal has a finite generating set by Hilbert basis theorem \cite{cox_ideals_2007}. For the ideal $I$ generated by the position equations, we define the associated variety as the set in Eq. \ref{eq:variety}.
\begin{eqnarray}\label{eq:variety}\nonumber
\mathbf{V}(I)=&\{&(\cis_a,\cis_b,\dots,\overline{\cis}_\alpha)\in\mathbb{C}^8:\\\nonumber
&{}&f_i(\cis_a,\cis_b,\dots,\overline{\cis}_\alpha)=0\\
&{}&,\forall f_i\in I\}
\end{eqnarray}
The Variety $\mathbf{V}(I)$ is the set of the solutions in the affine space $\mathbb{C}^8$. Thus, the problem of finding the posture of the 3RPR manipulator reduces to determining the variety $\mathbf{V}(I)$. Furthermore, Proposition \ref{prop:varideal}, a more general version of which is found in \cite{cox_ideals_2007}, gives us the ability to use the generators of $I$ directly in calculating the affine variety.
\begin{proposition}\label{prop:varideal}
Let $\mathbf{V}(f_1,f_2,\dots,f_8)$ be defined as an affine variety given by:
\begin{eqnarray*}
\mathbf{V}(f_1,\dots,f_8)&=&\{(\cis_a,\dots,\overline{\cis}_\alpha)\in\mathbb{C}^8:\\
&{}&\forall f_i(\cis_a,\dots,\overline{\cis}_\alpha)=0\}
\end{eqnarray*}
If $I=\left<f_1,f_2,\dots,f_8\right>$, then $\mathbf{V}(I)=\mathbf{V}(f_1,\dots,f_8)$.
\end{proposition}
\begin{proof}
We let $(\cis_a,\hdots\overline{\cis}_\alpha)\in\mathbf{V}(I)$. So, for $f_i\in I$, we use Eq. \ref{eq:idealgen} to write  $f=\sum_{i=1}^8h_if_i\in I$ with $h_i\in\mathbb{C}[\cis_a,\cis_b,\cis_c,\cis_\alpha,\overline{\cis}_a,\overline{\cis}_b,\overline{\cis}_c,\overline{\cis}_\alpha]$. Thus, $f(\cis_a,\hdots,\overline{\cis}_\alpha)=0$, and in particular, $f_i(\cis_a,\dots,\overline{\cis}_\alpha)=0$. Thus, $(\cis_a,\hdots,\overline{\cis}_\alpha)\in\mathbf{V}(f_1,\hdots,f_8)$, and hence,  $\mathbf{V}(I)\subseteq\mathbf{V}(f_1,\dots,f_8)$.

Conversely, let us assume  $(\cis_a,\dots,\overline{\cis}_\alpha)\in\mathbf{V}(f_1,\dots,f_8)$. Since $f_i(\cis_a,\dots,\overline{\cis}_\alpha)=0$, we write $f=\sum_{i=1}^8h_if_i=0$. 
Because $f_1,\dots,f_8\in I$, we write $f\in I$. Therefore, $(\cis_a,\dots,\overline{\cis}_\alpha)\in \mathbf{V}(I)$, and $\mathbf{V}(f_1,\dots,f_8)\subseteq\mathbf{V}(I)$.

Hence, because $\mathbf{V}(I)\subseteq\mathbf{V}(f_1,\dots,f_8)$ and $\mathbf{V}(f_1,\dots,f_8)\subseteq\mathbf{V}(I)$, we must have $\mathbf{V}(I)=\mathbf{V}(f_1,\dots,f_8)$.
\end{proof}

Thus, as a consequence of the forward position problem, we would like to find the set $\mathbf{V}(I)=\mathbf{V}(f_1,\dots,f_8)$ for given $f_i$ in the Proposition \ref{prop:ideal}. We can use Groebner bases theory to find $\mathbf{V}(I)$. The Groebner basis is a set of polynomials the leading terms of which generate the same ideal as that of the leading terms of the ideal. Consequently, a Groebner basis of the ideal $I$ is a basis of $I$. Thus, the solution of the Groebner basis would give us $\mathbf{V}(I)$. Additional information about Groebner bases theory and the associated Buchberger algorithm can be found in \cite{cox_ideals_2007}.

We compute the Groebner basis in the numerical examples by using Sage computer algebra system. The computations are performed in the field of Gaussian rational numbers ($\mathbb{Q}(\mi)$) with the following lex order $\cis_a\lex \cis_b\lex \cis_c\lex \cis_\alpha\lex \overline{\cis}_a\lex \overline{\cis}_b\lex \overline{\cis}_c\lex\overline{\cis}_\alpha$. The 3RPR in-parallel manipulator has a base and the moving platform in right triangular shapes. In particular, we use $l_{ac}=4$, $l_{ab}=3$, $d_{ac}=8\mi$, $d_{ab}=6$, and $\cis_\beta=\cis(\pi/2)=\mi$ as the numerical data in the examples. We substitute these values together with the connector lengths to each $f_i$ of the Proposition \ref{prop:ideal} for the solution.

We provide two examples for computations. In the first example, we use the numerical values for the connector lengths as $s_a=2$, $s_b=7/2$, and $s_c=5/2$. In the second numerical example, we use the connector lengths $s_a=7/2$, $s_b=6$, and $s_c=15/2$.
\subsection{Solving Position Equations of 3RPR In-Parallel Planar Manipulator}
We solve the polynomial equations by using elimination ideals. For $I=\left<f_1,\dots,f_8\right>$, we define the $l$-th elimination ideal $I_l$ as
\[
I_l=I\cap \mathbb{C}[\cis_{l+1},\dots,\overline{\cis}_\alpha]
\]
where $\mathbb{C}[\cis_{l+1},\dots,\overline{\cis}_\alpha]$ represents the remaining $8-l$ elements of $\cis_a,\dots,\overline{\cis}_\alpha$. Thus, the elimination ideal $I_l$ contains polynomials only in terms of $\cis_{l+1},\dots,\overline{\cis}_\alpha$, which are written according to lex order. To find the polynomials constituting the $I_l$, we use Groebner basis. For a basis $G$ associated with the ideal $I$, we write
\[
G_l=G\cap\mathbb{C}[\cis_{l+1},\dots,\overline{\cis}_\alpha]
\]
as the Groebner basis of the $l$-th elimination ideal $I_l$ \cite{cox_ideals_2007}.

For each of the $G_l$, we find the corresponding partial solution set, $\mathbf{V}(I_l)$. Then, we increase the variables by one and form the Groebner basis $G_{l-1}$ of the ideal $I_{l-1}$. We calculate the additional coordinate and stack to the partial solution. We repeat until we find all the coordinates.
\subsubsection{Solution Set for Example 1}\label{subsec:solex1}
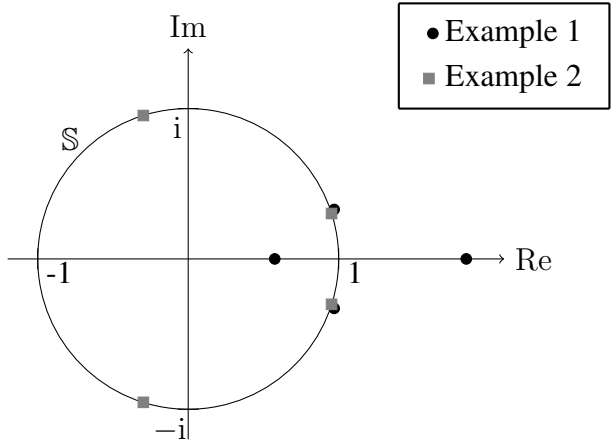
\begin{figure}[!t]
	\begin{center}
	 	\begin{tikzpicture}[scale=2]
	\draw[->] (-1.2,0) -- (2.1,0) node [right] {$\mathrm{Re}$};
	\draw[->] (0,-1.2) -- (0,1.4) node [above] {$\mathrm{Im}$};
	\draw (0,0) circle (1) node [above left=37pt] {$\mathbb{S}$};
	\foreach \x in {-1,1}
	\draw(\x,2pt) -- (\x,-2pt) node [below=1pt, right=-1pt] {\x};
	\draw(2pt,1) -- (-2pt,1) node [left, below=-1pt] {$\mathrm{i}$};
	\draw(2pt,-1) -- (-2pt,-1) node [left=3pt, below=-1pt] {$-\mathrm{i}$};
	\node[mark size=2pt,color=black] at (0.549,0) {\pgfuseplotmark{*}};
	\node[mark size=2pt,color=black] at (1.822,0) {\pgfuseplotmark{*}};
	\node[mark size=2pt,color=black] at (0.944,-0.329) {\pgfuseplotmark{*}};
	\node[mark size=2pt,color=black] at (0.944,0.329) {\pgfuseplotmark{*}};
	\node[mark size=2pt,color=gray] at (-0.298,-0.954) {\pgfuseplotmark{square*}};
	\node[mark size=2pt,color=gray] at (-0.298,0.954) {\pgfuseplotmark{square*}};
	\node[mark size=2pt,color=gray] at (0.953,-0.302) {\pgfuseplotmark{square*}};
	\node[mark size=2pt,color=gray] at (0.953,0.302) {\pgfuseplotmark{square*}};
	\draw[thick] (1.4,1) -- (2.8,1);
	\draw[thick] (2.8,1) -- (2.8,1.7);
	\draw[thick] (2.8,1.7) -- (1.4,1.7);
	\draw[thick] (1.4,1.7) -- (1.4,1);
	\node[mark size=2pt,color=black] at (1.6,1.5) {\pgfuseplotmark{*}};
	\node at (1.6,1.5) [right=2pt] {Example 1};
	\node[mark size=2pt,color=gray] at (1.6,1.2) {\pgfuseplotmark{square*}}; 
	\node at (1.6,1.2) [right=2pt] {Example 2};
	\end{tikzpicture}
	\end{center}
	\caption{Roots of the first elimination ideals for the examples with respect to the unit circle $\mathbb{S}$.}
	\label{fig:unit_circle-sol1}
\end{figure}

For the example with connector lengths $s_a=2$, $s_b=7/2$, and $s_c=5/2$, we compute the Groebner basis and include it in the Appendix \ref{app:a-G}. We inspect the polynomials $g_i$ of the Groebner basis in the first numerical example to find $I_7$ as:
\[
I_7=I\cap\mathbb{C}[\overline{\cis}_\alpha]=\left<g_8\right>
\]
with $g_8$ being taken from the Appendix \ref{app:a-G} as 
\begin{eqnarray}\label{eq:I7}\nonumber
g_8 &=& \overline{\cis}_\alpha^4-\frac{213}{50}\overline{\cis}_\alpha^3\\
&{}&+\frac{165857}{25600}\overline{\cis}_\alpha^2-\frac{213}{50}\overline{\cis}_\alpha+1
\end{eqnarray}
We note that the polynomial in Eq. \ref{eq:I7} is self reciprocal. Solving the equation $g_8=0$, we obtain the partial solution $\mathbf{V}(I_7)$. Consequently, we calculate:
\begin{eqnarray*}
\mathbf{V}(I_7)&=&\{\overline{\cis}_\alpha:0.549,1.822,0.944-0.329\mi\\&{}&,0.944+0.329\mi\}
\end{eqnarray*}
We draw the roots in Fig. \ref{fig:unit_circle-sol1} with respect to the unit circle. Thus, we find that two of the solutions are real while the other two are complex conjugates located on $\mathbb{S}$. Furthermore, we observe that the real roots form an inversive pair. By Proposition \ref{prop:harmonic_conj}, we conclude that the real roots are harmonic conjugates.

Discarding the two real solutions from $\mathbf{V}(I_7)$, we continue to extend the solution set to $\mathbf{V}(I_6)$. We note that \[I_6=I\cap\left\{\overline{\cis}_c,\overline{\cis}_\alpha\right\}=\left<g_7,g_8\right>\]
with $g_7$ taken from the Appendix \ref{app:a-G} as
\begin{eqnarray*}
g_7 &=& \overline{\cis}_c+\left(-\frac{14080}{2017}+\frac{2880}{2017}\mi\right)\overline{\cis}_\alpha^3\\\nonumber
&{}&+\left(\frac{264704}{10085}-\frac{54144}{10085}\mi\right)\overline{\cis}_\alpha^2\\\nonumber
&{}&+\left(-\frac{254163}{8063}+\frac{1188649}{161360}\mi\right)\overline{\cis}_\alpha\\
&{}&+\frac{128698}{10085}-\frac{44936}{10085}\mi
\end{eqnarray*}
Thus, we use $g_7=0$ to solve for the coordinates $\overline{\cis}_c$ to obtain
\begin{eqnarray*}
\mathbf{V}(I_6)&=&\{(\overline{\cis}_c,\overline{\cis}_\alpha):(-0.135+0.991\mi\\
&{}&,0.944-0.329\mi),(0.944+0.329\mi\\
&{}&,0.451+0.892\mi)\}
\end{eqnarray*}

We continue extending until we obtain the variety of the ideal $\mathbf{V}(I)$ as:
\begin{eqnarray}\nonumber
\mathbf{V}(I)&=&\{(\cis_a,\cis_b,\cis_c,\cis_\alpha,\overline{\cis}_a,\overline{\cis}_b\\\nonumber
&{}&,\overline{\cis}_c,\overline{\cis}_\alpha):(0.489+0.872\mi\\\nonumber
&{}&,-0.626+0.780\mi,-0.135-0.991\mi\\\nonumber
&{}&,0.944+0.329\mi,0.489-0.872\mi\\\nonumber
&{}&,-0.626-0.780\mi,-0.135+0.991\mi\\\nonumber
&{}&,0.944-0.329\mi),(-0.093+0.996\mi\\\nonumber
&{}&,-0.958+0.287\mi,0.451-0.892\mi\\\nonumber
&{}&,0.944-0.329\mi,-0.093-0.996\mi\\\nonumber
&{}&,-0.958-0.287\mi,0.451+0.892\mi\\
&{}&,0.944+0.329\mi)\}
\end{eqnarray}

\begin{figure*}[!t]
 \centering
\subfloat[$\theta_a=60.75$, $\theta_b=128.72$, $\theta_c=-97.75$, $\alpha=19.18$.]{\begin{tikzpicture}
[scale=0.70]
\draw[step=2,gray!50,very thin] (0,0) grid (6,8);
\draw(0,0) -- (60.75:1);
\draw(60.75:1) --++ (150.75:0.2) --++ (60.75:0.5);
\draw(60.75:1) --++ (-30.75:0.2) --++ (60.75:0.5);
\draw(60.75:1.2) -- (60.75:2);
\draw(60.75:1.2) --++ (150.75:0.1);
\draw(60.75:1.2) --++ (-30.75:0.1);
\draw[black!20,fill] (0.97724248,1.744992) -- (3.8107158,2.7306029) -- (-0.33690543,5.5229564) -- (0.97724248,1.744992);
\draw(60.75:2) --++ (19.18:3);
\draw(60.75:2) --++ (109.18:4);
\draw[fill=white] (0,0) circle (0.2) node (ja1) {};
\draw (0.4,-0.4) -- (0.4,0) arc (0:180:.4) (-0.4,0) -- (-0.4,-0.4) -- 
(0.4,-0.4) node [left=20pt] {$A$};
  \foreach \x in {-0.4, -0.2, 0, 0.2, 0.4}
    \draw (\x,-0.4) -- (\x-0.1,-0.5);
\draw[fill=white] (60.75:2) circle (0.2) node [below right=-1pt] (ja3) {$O_a$};
\begin{scope}
[shift={(6,0)}]
\draw(0,0) -- (128.72:1);
\draw(128.72:1) --++ (218.72:0.2) --++ (128.72:0.5);
\draw(128.72:1) --++ (38.72:0.2) --++ (128.72:0.5);
\draw(128.72:1.2) --++ (128.72:2.3);
\draw(128.72:1.2) --++ (218.72:0.1);
\draw(128.72:1.2) --++ (38.72:0.1);
\draw[fill=white] (0,0) circle (0.2) node (jb1) {};
\draw (0.4,-0.4) -- (0.4,0) arc (0:180:.4) (-0.4,0) -- (-0.4,-0.4) -- 
(0.4,-0.4) node [left=20pt] {$B$};
\foreach \x in {-0.4, -0.2, 0, 0.2, 0.4}
  \draw (\x,-0.4) -- (\x-0.1,-0.5);
\draw[fill=white] (128.72:3.5) circle (0.2) node [right=4pt](jb3){$O_b$};
\end{scope}
\begin{scope}
[shift={(0,8)}]
\draw(0,0) -- (-97.75:1);
\draw(-97.75:1) --++ (-7.75:0.2) --++ (-97.75:0.5);
\draw(-97.75:1) --++ (-187.75:0.2) --++ (-97.75:0.5);
\draw(-97.75:1.2) --++ (-97.75:1.3);
\draw(-97.75:1.2) --++ (-7.75:0.1);
\draw(-97.75:1.2) --++ (-187.75:0.1);
\begin{scope}[rotate=180]
\draw[fill=white] (0,0) circle (0.2) node (jc1) {};
\draw (0.4,-0.4) -- (0.4,0) arc (0:180:.4) (-0.4,0) -- (-0.4,-0.4) -- 
(0.4,-0.4) node [below left=1pt] {$C$};
\foreach \x in {-0.4, -0.2, 0, 0.2, 0.4}
  \draw (\x,-0.4) -- (\x-0.1,-0.5);
\end{scope}
\draw[fill=white] (-97.75:2.5) circle (0.2) node [right=4pt](jc3){$O_c$};
\end{scope}
\end{tikzpicture} %
\label{fig:first_case}}
\hfil
\subfloat[$\theta_a=95.32$, $\theta_b=163.3$, $\theta_c=-63.16$, $\alpha=-19.18$.]{\begin{tikzpicture}
[scale=0.70]
\draw[step=2,gray!50,very thin] (0,0) grid (6,8);
\draw(0,0) -- (95.33:1);
\draw(95.33:1) --++ (185.33:0.2) --++ (95.33:0.5);
\draw(95.33:1) --++ (5.33:0.2) --++ (95.33:0.5);
\draw(95.33:1.2) -- (95.33:2);
\draw(95.33:1.2) --++ (185.33:0.1);
\draw(95.33:1.2) --++ (5.33:0.1);
\draw[black!20,fill] (-0.18578387,1.9913524) -- (2.6476895,1.0057415) -- (1.128364,5.7693168);
\draw(95.33:2) --++ (-19.18:3);
\draw(95.33:2) --++ (70.82:4);
\draw[fill=white] (0,0) circle (0.2) node (ja1) {};
\draw (0.4,-0.4) -- (0.4,0) arc (0:180:.4) (-0.4,0) -- (-0.4,-0.4) -- 
(0.4,-0.4) node [left=20pt] {$A$};
  \foreach \x in {-0.4, -0.2, 0, 0.2, 0.4}
    \draw (\x,-0.4) -- (\x-0.1,-0.5);
\draw[fill=white] (95.33:2) circle (0.2) node [below left=-1pt] (ja3) {$O_a$};
\begin{scope}
[shift={(6,0)}]
\draw(0,0) -- (163.30:1);
\draw(163.30:1) --++ (253.30:0.2) --++ (163.30:0.5);
\draw(163.30:1) --++ (73.30:0.2) --++ (163.30:0.5);
\draw(163.30:1.2) --++ (163.30:2.3);
\draw(163.30:1.2) --++ (253.30:0.1);
\draw(163.30:1.2) --++ (73.30:0.1);
\draw[fill=white] (0,0) circle (0.2) node (jb1) {};
\draw (0.4,-0.4) -- (0.4,0) arc (0:180:.4) (-0.4,0) -- (-0.4,-0.4) -- 
(0.4,-0.4) node [left=20pt] {$B$};
\foreach \x in {-0.4, -0.2, 0, 0.2, 0.4}
  \draw (\x,-0.4) -- (\x-0.1,-0.5);
\draw[fill=white] (163.30:3.5) circle (0.2) node [below left=-1pt](jb3){$O_b$};
\end{scope}
\begin{scope}
[shift={(0,8)}]
\draw(0,0) -- (-63.16:1);
\draw(-63.16:1) --++ (26.84:0.2) --++ (-63.16:0.5);
\draw(-63.16:1) --++ (-153.16:0.2) --++ (-63.16:0.5);
\draw(-63.16:1.2) --++ (-63.16:1.3);
\draw(-63.16:1.2) --++ (26.84:0.1);
\draw(-63.16:1.2) --++ (-153.16:0.1);
\begin{scope}
[rotate=180]
\draw[fill=white] (0,0) circle (0.2) node (jc1) {};
\draw (0.4,-0.4) -- (0.4,0) arc (0:180:.4) (-0.4,0) -- (-0.4,-0.4) -- 
(0.4,-0.4) node [below left=1pt] {$C$};
\foreach \x in {-0.4, -0.2, 0, 0.2, 0.4}
  \draw (\x,-0.4) -- (\x-0.1,-0.5);
\end{scope}
\draw[fill=white] (-63.16:2.5) circle (0.2) node [right=4pt](jc3){$O_c$};
\end{scope}
\end{tikzpicture}%
\label{fig:second_case}}
\caption{Assembly configurations of the solution in Eq. \ref{eq:solex1}, example \ref{subsec:solex1}.}
\label{fig:ex1}
\end{figure*}
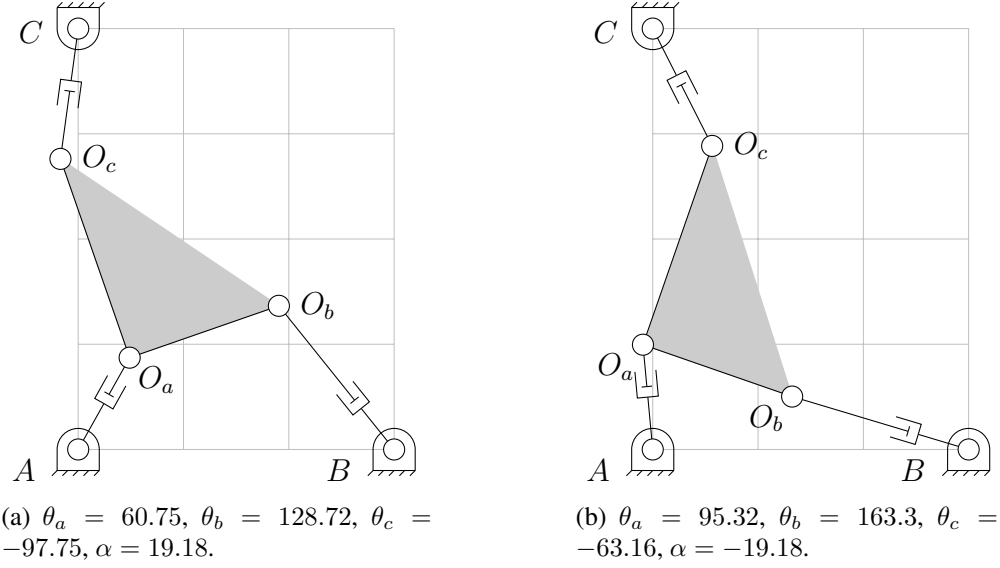

We convert the solutions of $\mathbf{V}(I)$ to angular quantities by using Eq. 6 available in \cite{2015arXiv150801733S} to obtain the solutions in degrees as given in Eq. \ref{eq:solex1}.
\begin{eqnarray}\nonumber
\label{eq:solex1}
S &=& \{(\theta_a,\theta_b,\theta_c,\alpha):(60.75,128.72,-97.75\\\nonumber
&{}&,19.18),(95.32,163.3,-63.16\\
&{}&,-19.18)\}
\end{eqnarray}
The postures of the in-parallel planar manipulator corresponding to the solution set $S$ is given in Fig. \ref{fig:ex1}.
\subsubsection{Solution Set for Example 2} \label{subsec:solex2}
For the connector lengths $s_a=7/2$, $s_b=6$, and $s_c=15/2$, we compute the Groebner bases, which we include in the Appendix \ref{app:b-H}. We inspect the polynomials of the Groebner bases, thus we find $I_7$ as:
\[
I_7 = I\cap\mathbb{C}[\overline{\cis}_\alpha]=\left<h_8\right>
\]
with $h_8$ being a self reciprocal polynomial, which is taken from the Appendix \ref{app:b-H} as
\begin{equation*}
h_8 = \overline{\cis}_\alpha^4-\frac{131}{100}\overline{\cis}_\alpha^3+\frac{12409}{14400}\overline{\cis}_\alpha^2-\frac{131}{100}\overline{\cis}_\alpha+1
\end{equation*}
Thus, the partial solution is obtained by solving $h_8=0$. Consequently, we obtain the roots as two pairs of complex conjugates given in Eq. \ref{eq:roots_ex2}.
\begin{eqnarray}\label{eq:roots_ex2}\nonumber
\mathbf{V}(I_7)&=&\{\overline{\cis}_\alpha:-0.298-0.954\mi,-0.298\\\nonumber
&{}&+0.954\mi,0.953-0.302\mi\\
&{}&,0.953+0.302\mi\}
\end{eqnarray}
We draw the roots given in the Eq. \ref{eq:roots_ex2} in Fig. \ref{fig:unit_circle-sol1} as gray squares. We observe that all the roots lie on the unit circle.

Proceeding as in the previous example and solving coordinates one at a time, we obtain the variety $\mathbf{V}(I)$ as:
\begin{eqnarray}\nonumber
\mathbf{V}(I)&=&\{(\cis_a,\cis_b,\cis_c,\cis_\alpha,\overline{\cis}_a,\overline{\cis}_b\\\nonumber
&{}&,\overline{\cis}_c,\overline{\cis}_\alpha):(0.869+0.495\mi\\\nonumber
&{}&,-0.642+0.766\mi,-0.104-0.995\mi\\\nonumber
&{}&,-0.298+0.954\mi,0.869-0.495\mi\\\nonumber
&{}&,-0.642-0.766\mi,-0.104+0.995\mi\\\nonumber
&{}&,-0.298-0.954),(0.262+0.965\mi\\\nonumber
&{}&,-0.996-0.086\mi,0.631-0.775\mi\\\nonumber
&{}&,-0.298-0.954\mi,0.262-0.965\mi\\\nonumber
&{}&,-0.996-0.086\mi,0.631+0.775\mi\\\nonumber
&{}&,-0.298+0.954\mi),(-0.776-0.830\mi\\\nonumber
&{}&,-0.976-0.217\mi,-0.523-0.852\mi\\\nonumber
&{}&,0.954+0.301\mi,-0.776+0.63\mi\\\nonumber
&{}&,-0.976+0.217\mi,-0.523+0.852\mi\\\nonumber
&{}&,0.953-0.302\mi),(-0.322-0.947\mi\\\nonumber
&{}&,-0.711-0.703\mi,0.01-\mi\\\nonumber
&{}&,0.954-0.301\mi,-0.322+0.947\mi\\\nonumber
&{}&,-0.711+0.703\mi,0.01+\mi\\
&{}&,0.953+0.302\mi)\}
\end{eqnarray}

\begin{figure*}[!t]
\centering
\subfloat[$\theta_a=29.7$, $\theta_b=129.98$, $\theta_c=-95.95$, $\alpha-107.36$.]{\begin{tikzpicture}
[scale=0.62]
\draw[step=2,gray!50,very thin] (0,0) grid (6,8);
\draw(0,0) -- (29.70:1);
\draw(29.70:1) --++ (119.70:0.2) --++ (29.70:0.5);
\draw(29.70:1) --++ (-60.30:0.2) --++ (29.70:0.5);
\draw(29.30:1.2) --++ (29.30:2.3);
\draw(29.70:1.2) --++ (119.70:0.1);
\draw(29.70:1.2) --++ (-60.30:0.1);
\draw[black!20,fill] (3.0402103,1.7341053) -- (2.1450867,4.5974519) -- (-0.77758516,0.5406072) -- (3.0402103,1.7341053);
\draw(29.70:3.5) --++ (107.36:3);
\draw(29.70:3.5) --++ (197.36:4);
\draw[fill=white] (0,0) circle (0.2) node (ja1) {};
\draw (0.4,-0.4) -- (0.4,0) arc (0:180:.4) (-0.4,0) -- (-0.4,-0.4) -- 
(0.4,-0.4) node [left=20pt] {$A$};
  \foreach \x in {-0.4, -0.2, 0, 0.2, 0.4}
    \draw (\x,-0.4) -- (\x-0.1,-0.5);
\draw[fill=white] (29.70:3.5) circle (0.2) node [above left=-1pt] (ja3) {$O_a$};
\begin{scope}
[shift={(6,0)}]
\draw(0,0) -- (129.98:1);
\draw(129.98:1) --++ (219.98:0.2) --++ (129.98:0.5);
\draw(129.98:1) --++ (39.98:0.2) --++ (129.98:0.5);
\draw(129.98:1.2) --++ (129.98:4.8);
\draw(129.98:1.2) --++ (219.98:0.1);
\draw(129.98:1.2) --++ (39.98:0.1);
\draw[fill=white] (0,0) circle (0.2) node (jb1) {};
\draw (0.4,-0.4) -- (0.4,0) arc (0:180:.4) (-0.4,0) -- (-0.4,-0.4) -- 
(0.4,-0.4) node [left=20pt] {$B$};
\foreach \x in {-0.4, -0.2, 0, 0.2, 0.4}
  \draw (\x,-0.4) -- (\x-0.1,-0.5);
\draw[fill=white] (129.98:6) circle (0.2) node [right=4pt](jb3){$O_b$};
\end{scope}
\begin{scope}
[shift={(0,8)}]
\draw(0,0) -- (-95.95:1);
\draw(-95.95:1) --++ (-5.95:0.2) --++ (-95.95:0.5);
\draw(-95.95:1) --++ (-185.95:0.2) --++ (-95.95:0.5);
\draw(-95.95:1.2) --++ (-95.95:6.3);
\draw(-95.95:1.2) --++ (-5.95:0.1);
\draw(-95.95:1.2) --++ (-185.95:0.1);
\begin{scope}[rotate=180]
\draw[fill=white] (0,0) circle (0.2) node (jc1) {};
\draw (0.4,-0.4) -- (0.4,0) arc (0:180:.4) (-0.4,0) -- (-0.4,-0.4) -- 
(0.4,-0.4) node [below left=1pt] {$C$};
\foreach \x in {-0.4, -0.2, 0, 0.2, 0.4}
  \draw (\x,-0.4) -- (\x-0.1,-0.5);
\end{scope}
\draw[fill=white] (-95.95:7.5) circle (0.2) node [left=4pt](jc3){$O_c$};
\end{scope}
\end{tikzpicture} %
\label{fig_first_case}}
\hfil
\subfloat[$\theta_a=74.8$, $\theta_b=175.08$, $\theta_c=-50.85$, $\alpha=-107.36$. ]{\begin{tikzpicture}
[scale=0.62]
\draw[step=2,gray!50,very thin] (0,0) grid (6,8);
\draw(0,0) -- (74.80:1);
\draw(74.80:1) --++ (164.80:0.2) --++ (74.80:0.5);
\draw(74.80:1) --++ (-15.2:0.2) --++ (74.2:0.5);
\draw(74.80:1.2) --++ (74.80:2.3);
\draw(74.80:1.2) --++ (164.80:0.1);
\draw(74.80:1.2) --++ (-15.2:0.1);
\draw[black!20,fill] (0.91766213,3.3775577) -- (0.022538523,0.51421113) -- (4.7354576,2.1840596) -- (0.91766213,3.3775577);
\draw(74.80:3.5) --++ (-107.36:3);
\draw(74.80:3.5) --++ (-17.36:4);
\draw[fill=white] (0,0) circle (0.2) node (ja1) {};
\draw (0.4,-0.4) -- (0.4,0) arc (0:180:.4) (-0.4,0) -- (-0.4,-0.4) -- 
(0.4,-0.4) node [left=20pt] {$A$};
  \foreach \x in {-0.4, -0.2, 0, 0.2, 0.4}
    \draw (\x,-0.4) -- (\x-0.1,-0.5);
\draw[fill=white] (74.80:3.5) circle (0.2) node [above left=-1pt] (ja3) {$O_a$};
\begin{scope}
[shift={(6,0)}]
\draw(0,0) -- (175.08:1);
\draw(175.08:1) --++ (265.08:0.2) --++ (175.08:0.5);
\draw(175.08:1) --++ (85.08:0.2) --++ (175.08:0.5);
\draw(175.08:1.2) --++ (175.08:4.8);
\draw(175.08:1.2) --++ (265.08:0.1);
\draw(175.08:1.2) --++ (85.08:0.1);
\draw[fill=white] (0,0) circle (0.2) node (jb1) {};
\draw (0.4,-0.4) -- (0.4,0) arc (0:180:.4) (-0.4,0) -- (-0.4,-0.4) -- 
(0.4,-0.4) node [left=20pt] {$B$};
\foreach \x in {-0.4, -0.2, 0, 0.2, 0.4}
  \draw (\x,-0.4) -- (\x-0.1,-0.5);
\draw[fill=white] (175.08:6) circle (0.2) node [left=4pt](jb3){$O_b$};
\end{scope}
\begin{scope}
[shift={(0,8)}]
\draw(0,0) -- (-50.85:1);
\draw(-50.85:1) --++ (39.15:0.2) --++ (-50.85:0.5);
\draw(-50.85:1) --++ (-140.85:0.2) --++ (-50.85:0.5);
\draw(-50.85:1.2) --++ (-50.85:6.3);
\draw(-50.85:1.2) --++ (39.15:0.1);
\draw(-50.85:1.2) --++ (-140.85:0.1);
\begin{scope}[rotate=180]
\draw[fill=white] (0,0) circle (0.2) node (jc1) {};
\draw (0.4,-0.4) -- (0.4,0) arc (0:180:.4) (-0.4,0) -- (-0.4,-0.4) -- 
(0.4,-0.4) node [below left=1pt] {$C$};
\foreach \x in {-0.4, -0.2, 0, 0.2, 0.4}
  \draw (\x,-0.4) -- (\x-0.1,-0.5);
\end{scope}
\draw[fill=white] (-50.85:7.5) circle (0.2) node [right=4pt](jc3){$O_c$};
\end{scope}
\end{tikzpicture}%
\label{fig_second_case}}
\hfil
\subfloat[$\theta_a=-140.94$, $\theta_b=-167.48$, $\theta_c=-121.54$, $\alpha=17.55$]{  \begin{tikzpicture}
  [scale=0.62]
  	\draw[step=2,gray!50,very thin] (-4,-2) grid (6,8);
    \draw(0,0) -- (-140.94:1);
    \draw(-140.94:1) --++ (-50.94:0.2) --++ (-140.94:0.5);
    \draw(-140.94:1) --++ (-230.94:0.2) --++ (-140.94:0.5);
    \draw(-140.94:1.2) --++ (-140.94:2.3);
    \draw(-140.94:1.2) --++ (-50.94:0.1);
    \draw(-140.94:1.2) --++ (-230.94:0.1);
    \draw[black!20,fill] (-2.7177028,-2.2054685) -- (0.14265972,-1.3008547) -- (-3.9238546,1.6083481) -- (-2.7177028,-2.2054685);
    \draw(-140.94:3.5) --++ (17.55:3);
    \draw(-140.94:3.5) --++ (107.55:4);
    \draw[fill=white] (0,0) circle (0.2) node (ja1) {};
    \draw (0.4,-0.4) -- (0.4,0) arc (0:180:.4) (-0.4,0) -- (-0.4,-0.4) -- 
(0.4,-0.4) node [above left=14pt] {$A$};
    \foreach \x in {-0.4, -0.2, 0, 0.2, 0.4}
      \draw (\x,-0.4) -- (\x-0.1,-0.5);
    \draw[fill=white] (-140.94:3.5) circle (0.2) node [below left=-1pt] (ja3) {$O_a$};
    \begin{scope}
    [shift={(6,0)}]
      \draw(0,0) -- (-167.48:1);
      \draw(-167.48:1) --++ (-77.48:0.2) --++ (-167.48:0.5);
      \draw(-167.48:1) --++ (-257.48:0.2) --++ (-167.48:0.5);
      \draw(-167.48:1.2) --++ (-167.48:4.8);
      \draw(-167.48:1.2) --++ (-77.48:0.1);
      \draw(-167.48:1.2) --++ (-257.48:0.1);
      \draw[fill=white] (0,0) circle (0.2) node (jb1) {};
      \draw (0.4,-0.4) -- (0.4,0) arc (0:180:.4) (-0.4,0) -- (-0.4,-0.4) -- (0.4,-0.4) node [above left=14pt] {$B$};
      \foreach \x in {-0.4, -0.2, 0, 0.2, 0.4}
        \draw (\x,-0.4) -- (\x-0.1,-0.5);
      \draw[fill=white] (-167.48:6) circle (0.2) node [below right=0pt](jb3){$O_b$};
    \end{scope}
    \begin{scope}
    [shift={(0,8)}]
      \draw(0,0) -- (-121.54:1);
      \draw(-121.54:1) --++ (-31.54:0.2) --++ (-121.54:0.5);
      \draw(-121.54:1) --++ (-211.54:0.2) --++ (-121.54:0.5);
      \draw(-121.54:1.2) --++ (-121.54:6.3);
      \draw(-121.54:1.2) --++ (-31.54:0.1);
      \draw(-121.54:1.2) --++ (-211.54:0.1);
      \begin{scope}[rotate=180]
      \draw[fill=white] (0,0) circle (0.2) node (jc1) {};
      \draw (0.4,-0.4) -- (0.4,0) arc (0:180:.4) (-0.4,0) -- (-0.4,-0.4) -- 
      (0.4,-0.4) node [below right=14pt] {$C$};
      \foreach \x in {-0.4, -0.2, 0, 0.2, 0.4}
        \draw (\x,-0.4) -- (\x-0.1,-0.5);
      \end{scope}
      \draw[fill=white] (-121.54:7.5) circle (0.2) node [left=4pt](jc3){$O_c$};
    \end{scope}
  \end{tikzpicture}%
\label{fig_third_case}}
\hfil
\subfloat[$\theta_a=-108.79$, $\theta_b=-135.33$, $\theta_c=-89.4$, $\alpha=-17.55$.]{  \begin{tikzpicture}
  [scale=0.62]
  	\draw[step=2,gray!50,very thin] (-2,-4) grid (6,8);
    \draw(0,0) -- (-108.79:1);
    \draw(-108.79:1) --++ (-18.79:0.2) --++ (-108.79:0.5);
    \draw(-108.79:1) --++ (-198.79:0.2) --++ (-108.79:0.5);
    \draw(-108.79:1.2) --++ (-108.79:2.3);
    \draw(-108.79:1.2) --++ (-18.79:0.1);
    \draw(-108.79:1.2) --++ (-198.79:0.1);
    \draw[fill=white] (0,0) circle (0.2) node (ja1) {};
    \draw (0.4,-0.4) -- (0.4,0) arc (0:180:.4) (-0.4,0) -- (-0.4,-0.4) -- 
(0.4,-0.4) node [above left=14pt] {$A$};
    \foreach \x in {-0.4, -0.2, 0, 0.2, 0.4}
      \draw (\x,-0.4) -- (\x-0.1,-0.5);
        \draw[black!20,fill] (-1.1273516,-3.3134692) -- (1.7330109,-4.2180831) -- (0.078800199,0.50034747) -- (-1.1273516,-3.3134692);
        \draw(-108.79:3.5) --++ (-17.55:3);
        \draw(-108.79:3.5) --++ (72.45:4);
        \draw[fill=white] (-108.79:3.5) circle (0.2) node [above left=-1pt] (ja3) {$O_a$};
    \begin{scope}
    [shift={(6,0)}]
      \draw(0,0) -- (-135.33:1);
      \draw(-135.33:1) --++ (-45.33:0.2) --++ (-135.33:0.5);
      \draw(-135.33:1) --++ (-225.33:0.2) --++ (-135.33:0.5);
      \draw(-135.33:1.2) --++ (-135.33:4.8);
      \draw(-135.33:1.2) --++ (-45.33:0.1);
      \draw(-135.33:1.2) --++ (-225.33:0.1);
      \draw[fill=white] (0,0) circle (0.2) node (jb1) {};
      \draw (0.4,-0.4) -- (0.4,0) arc (0:180:.4) (-0.4,0) -- (-0.4,-0.4) -- (0.4,-0.4) node [above left=14pt] {$B$};
      \foreach \x in {-0.4, -0.2, 0, 0.2, 0.4}
        \draw (\x,-0.4) -- (\x-0.1,-0.5);
      \draw[fill=white] (-135.33:6) circle (0.2) node [below right=0pt](jb3){$O_b$};
    \end{scope}
    \begin{scope}
    [shift={(0,8)}]
      \draw(0,0) -- (-89.40:1);
      \draw(-89.40:1) --++ (0.60:0.2) --++ (-89.40:0.5);
      \draw(-89.40:1) --++ (-179.40:0.2) --++ (-89.40:0.5);
      \draw(-89.40:1.2) --++ (-89.40:6.3);
      \draw(-89.40:1.2) --++ (0.60:0.1);
      \draw(-89.40:1.2) --++ (-179.40:0.1);
      \begin{scope}[rotate=180]
      \draw[fill=white] (0,0) circle (0.2) node (jc1) {};
      \draw (0.4,-0.4) -- (0.4,0) arc (0:180:.4) (-0.4,0) -- (-0.4,-0.4) -- 
      (0.4,-0.4) node [below right=14pt] {$C$};
      \foreach \x in {-0.4, -0.2, 0, 0.2, 0.4}
        \draw (\x,-0.4) -- (\x-0.1,-0.5);
      \end{scope}
      \draw[fill=white] (-89.40:7.5) circle (0.2) node [right=4pt](jc3){$O_c$};
    \end{scope}
  \end{tikzpicture}%
\label{fig_fourth_case}}
\caption{Assembly configurations corresponding to the solution in Eq. \ref{eq:solex2}, example \ref{subsec:solex2}.}
\label{fig:ex2}
\end{figure*}
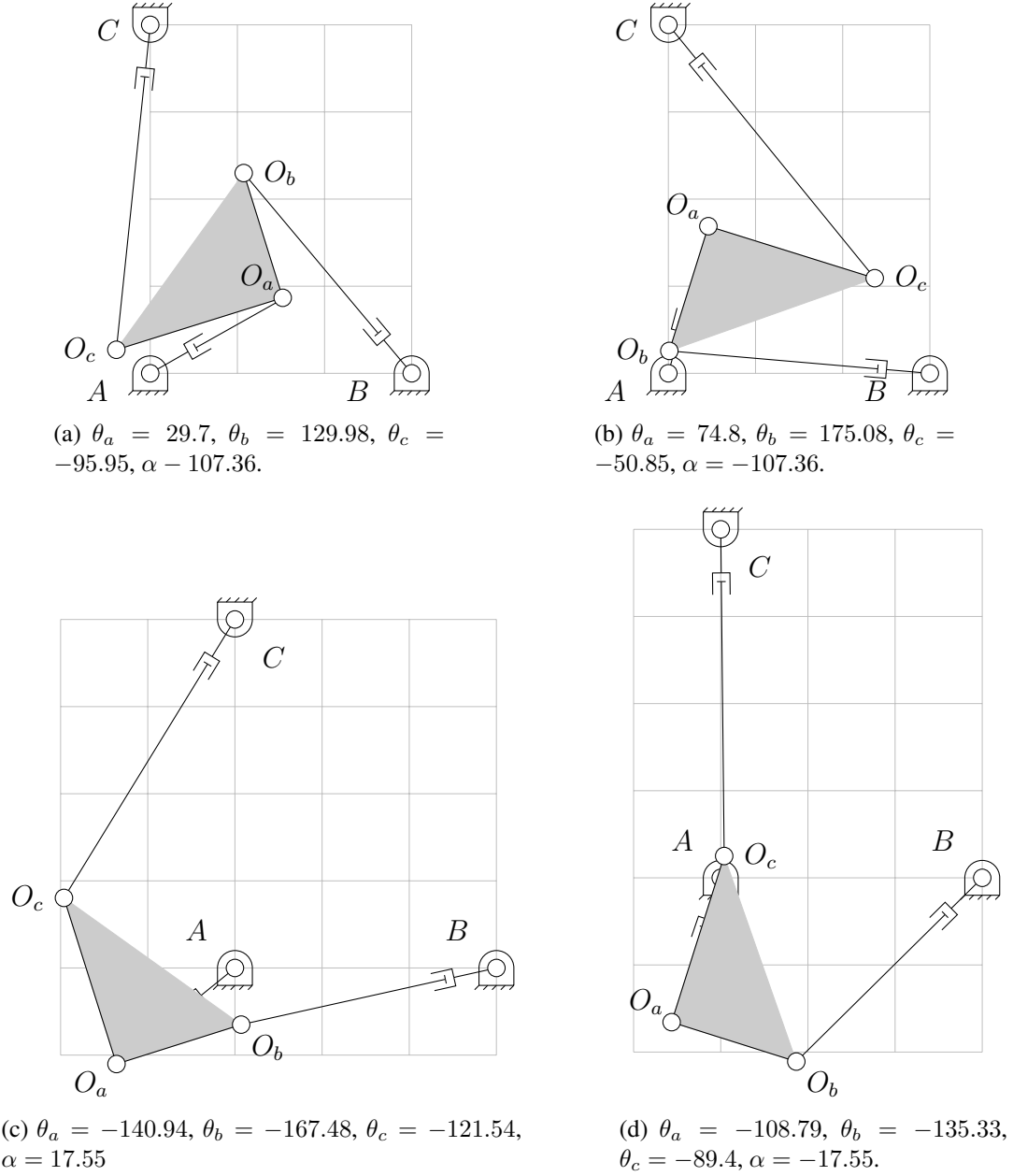

We transform the solution $\mathbf{V}(I)$ to $\mathbb{S}^4$ by the same procedure as in the Example 1 to obtain the solution set in degrees in Eq. \ref{eq:solex2}.
\begin{eqnarray}\label{eq:solex2}\nonumber
S&=&\{(\theta_a,\theta_b,\theta_c,\alpha):(29.7,129.98,-95.95\\\nonumber
&{}&,107.36),(74.8,175.08,-50.85\\\nonumber
&{}&,-107.36),(-140.94,-167.48,-121.54\\\nonumber
&{}&,17.55),(-108.79,-135.33,-89.4\\
&{}&,-17.55)\}
\end{eqnarray}
We draw the posture of the 3RPR in-parallel manipulator by using the solution set $S$ in Eq. \ref{eq:solex2} as shown in Fig. \ref{fig:ex2}.


\section{Conclusion}
In this paper, we solve the position equations of a 3RPR in-parallel planar manipulator, which are essentially linear equations with the constraints associated with revolute joints, in the Gauss plane  in order to obtain the posture of the moving platform. We use Groebner bases theory to obtain the solution of the resulting ideal consisting of four planes and circles. As can be seen from the concrete examples for the 3RPR manipulator, the elimination ideal which contain only the single variable polynomial are used as the base step for obtaining the solutions. The single variable polynomials are self reciprocal in $\mathbb{Q}(\mi)$, thus we obtain either points on the circle or on the real axis. Thus, we obtain either harmonic conjugates on the real line or complex conjugates on the unit circle. This behavior of the polynomials in solving the forward position analysis of the manipulator is not observed in other studies using $\mathbb{R}$ in the literature, or which use numerical methods.

Thus, as a conclusion, we emphasize the importance of working in the field $\mathbb{C}$, which is algebraically closed. While parametrization of spatial manipulators does not appear to be convenient in $\mathbb{C}$, this fact allows us to inspect position analysis of planar manipulators in more detail in $\mathbb{C}$ by using tools from algebra and geometry.

\begin{appendices}
  \section{Elements of the Groebner Basis $G$}\label{app:a-G}
We tabulate the elements of the Groebner basis of the example in Section \ref{subsec:solex1} below:
\begin{eqnarray*}
g_1 & = & \cis_a+\left(\frac{11549}{8068}-\frac{1117}{2017}\mi\right)\overline{\cis}_\alpha^3\\
&{}&+\left(-\frac{699937}{403400}+\frac{327921}{100850}\mi\right)\overline{\cis}_\alpha^2\\
&{}&+\left(-\frac{121048707}{206540800}-\frac{289403069}{51635200}\mi\right)\overline{\cis}_\alpha\\
&{}&+\frac{1529727}{1613600}+\frac{13606769}{6454400}\mi
\end{eqnarray*}
\begin{eqnarray*}
g_2& = & \cis_b+\left(\frac{23651}{14119}-\frac{4468}{14119}\mi\right)\overline{\cis}_\alpha^3\\
&{}&+\left(-\frac{3277663}{705950}+\frac{655842}{352975}\mi\right)\overline{\cis}_\alpha^2\\
&{}&+\left(\frac{1886152707}{361446400}-\frac{289403069}{90361600}\mi\right)\overline{\cis}_\alpha\\
&{}&-\frac{562911}{403400}+\frac{1942967}{1613600}\mi
\end{eqnarray*}
\begin{eqnarray*}
g_3 & = & \cis_c+\left(\frac{11549}{10085}+\frac{11668}{10085}\mi\right)\overline{\cis}_\alpha^3\\
&{}&+\left(-\frac{699937}{504250}-\frac{1062642}{252125}\mi\right)\overline{\cis}_\alpha^2\\
&{}&+\left(-\frac{121048707}{258176000}+\frac{379664069}{64544000}\mi\right)\overline{\cis}_\alpha\\
&{}&+\frac{1529727}{2017000}-\frac{15573119}{8068000}\mi
\end{eqnarray*}
\begin{equation*}
g_4=\cis_\alpha+\overline{\cis}_\alpha^3-\frac{213}{50}\overline{\cis}_\alpha^2+\frac{165857}{25600}\overline{\cis}_\alpha-\frac{213}{50}
\end{equation*}
\begin{eqnarray*}
g_5 &=& \overline{\cis}_a+\left(-\frac{17600}{2017}+\frac{3600}{2017}\mi\right)\overline{\cis}_\alpha^3\\
&{}&+\left(\frac{66176}{2017}-\frac{13536}{2017}\mi\right)\overline{\cis}_\alpha^2\\
&{}&+\left(-\frac{1270815}{32272}+\frac{930473}{129088}\mi\right)\overline{\cis}_\alpha\\
&{}&+\frac{64349}{4034}-\frac{3166}{2017}\mi
\end{eqnarray*}
\begin{eqnarray*}
g_6 &=& \overline{\cis}_b+\left(-\frac{70400}{14119}+\frac{14400}{14119}\mi\right)\overline{\cis}_\alpha^3\\
&{}&+\left(\frac{264704}{14119}-\frac{54144}{14119}\mi\right)\overline{\cis}_\alpha^2\\
&{}&+\left(-\frac{1319223}{56476}+\frac{930473}{225904}\mi\right)\overline{\cis}_\alpha\\
&{}&+\frac{152902}{14119}-\frac{12664}{14119}\mi
\end{eqnarray*}
\begin{eqnarray*}
g_7 &=& \overline{\cis}_c+\left(-\frac{14080}{2017}+\frac{2880}{2017}\mi\right)\overline{\cis}_\alpha^3\\
&{}&+\left(\frac{264704}{10085}-\frac{54144}{10085}\mi\right)\overline{\cis}_\alpha^2\\
&{}&+\left(-\frac{254163}{8063}+\frac{1188649}{161360}\mi\right)\overline{\cis}_\alpha\\
&{}&+\frac{128698}{10085}-\frac{44936}{10085}\mi
\end{eqnarray*}
\begin{equation*}
g_8 = \overline{\cis}_\alpha^4-\frac{213}{50}\overline{\cis}_\alpha^3+\frac{165857}{25600}\overline{\cis}_\alpha^2-\frac{213}{50}\overline{\cis}_\alpha+1
\end{equation*}


  \section{Elements of the Groebner Basis $H$}\label{app:b-H}
We tabulate the elements of the Groebner basis of the example in Section \ref{subsec:solex2}.
\begin{eqnarray*}
h_1 &=& \cis_a+\left(-\frac{50847}{185143}-\frac{66196}{185143}\mi\right)\overline{\cis}_\alpha^3\\
&{}&+\left(\frac{1765851}{2644900}+\frac{4147919}{4628575}\mi\right)\overline{\cis}_\alpha^2\\
&{}&+\left(-\frac{21536141}{888686400}-\frac{1231963}{95216400}\mi\right)\overline{\cis}_\alpha\\
&{}&+\frac{3059531}{15869400}+\frac{2633513}{9257150}\mi
\end{eqnarray*}
\begin{eqnarray*}
h_2 &=& \cis_b+\left(\frac{35949}{105796}-\frac{16549}{79347}\mi\right)\overline{\cis}_\alpha^3\\
&{}&+\left(-\frac{2809319}{10579600}+\frac{4147919}{7934700}\mi\right)\overline{\cis}_\alpha^2\\
&{}&+\left(\frac{211625047}{507820800}-\frac{8623741}{1142596800}\mi\right)\overline{\cis}_\alpha\\
&{}&+\frac{87116033}{190432800}+\frac{2633513}{15869400}\mi
\end{eqnarray*}
\begin{eqnarray*}
h_3 &=& \cis_c+\left(-\frac{16949}{132245}+\frac{145396}{396735}\mi\right)\overline{\cis}_\alpha^3\\
&{}&+\left(\frac{4120319}{13224500}-\frac{2781719}{9918375}\mi\right)\overline{\cis}_\alpha^2\\
&{}&+\left(-\frac{21536141}{1904328000}+\frac{647787541}{1428246000}\mi\right)\overline{\cis}_\alpha\\
&{}&+\frac{21416717}{238041000}+\frac{9933437}{19836750}\mi
\end{eqnarray*}
\begin{equation*}
h_4 = \cis_\alpha+\overline{\cis}_\alpha^3-\frac{131}{100}\overline{\cis}_\alpha^2+\frac{12409}{14400}\overline{\cis}_\alpha-\frac{131}{100}
\end{equation*}
\begin{eqnarray*}
h_5 &=& \overline{\cis}_a+\left(-\frac{114000}{185143}+\frac{158400}{185143}\mi\right)\overline{\cis}_\alpha^3\\
&{}&+\left(\frac{92340}{185143}-\frac{128304}{185143}\mi\right)\overline{\cis}_\alpha^2\\
&{}&+\left(\frac{327349}{2221716}-\frac{33449}{185143}\mi\right)\overline{\cis}_\alpha\\
&{}&+\frac{69306}{185143}-\frac{105208}{185143}\mi
\end{eqnarray*}
\begin{eqnarray*}
h_6 &=& \overline{\cis}_b+\left(-\frac{9500}{26449}+\frac{13200}{26449}\mi\right)\overline{\cis}_\alpha^3\\
&{}&+\left(\frac{7695}{26449}-\frac{10692}{26449}\mi\right)\overline{\cis}_\alpha^2\\
&{}&+\left(-\frac{1576979}{3808656}-\frac{33449}{317388}\mi\right)\overline{\cis}_\alpha\\
&{}&+\frac{64449}{52898}-\frac{26302}{79347}\mi
\end{eqnarray*}
\begin{eqnarray*}
h_7 &=& \overline{\cis}_c+\left(-\frac{7600}{26449}+\frac{10560}{26449}\mi\right)\overline{\cis}_\alpha^3\\
&{}&+\left(\frac{6156}{26449}-\frac{42768}{132245}\mi\right)\overline{\cis}_\alpha^2\\
&{}&+\left(\frac{327349}{4760820}+\frac{59381}{132245}\mi\right)\overline{\cis}_\alpha\\
&{}&+\frac{23102}{132245}-\frac{528392}{396735}\mi
\end{eqnarray*}
\begin{equation*}
h_8 = \overline{\cis}_\alpha^4-\frac{131}{100}\overline{\cis}_\alpha^3+\frac{12409}{14400}\overline{\cis}_\alpha^2-\frac{131}{100}\overline{\cis}_\alpha+1
\end{equation*}


\end{appendices}
\bibliography{manuscript}
\end{document}